\documentclass[10pt,twocolumn,letterpaper]{article}

\usepackage{cvpr}              

\usepackage{graphicx}
\usepackage{amsmath}
\usepackage{amssymb}
\usepackage{booktabs}

\usepackage[pagebackref,breaklinks,colorlinks]{hyperref}

\usepackage[capitalize]{cleveref}
\crefname{section}{Sec.}{Secs.}
\Crefname{section}{Section}{Sections}
\Crefname{table}{Table}{Tables}
\crefname{table}{Tab.}{Tabs.}

\def\cvprPaperID{7} 
\def\confName{CVPR}
\def\confYear{2023}

\usepackage{multirow} 

\begin{document}

\title{Open-TransMind: A New Baseline and Benchmark for 1st Foundation Model Challenge of Intelligent Transportation}

\author
{ 
Yifeng Shi\textsuperscript{1}\thanks{indicates equal contribution}\thanks{Yifeng Shi(shiyifeng@baidu.com is the corresponding author.)}
\quad
Feng Lv\textsuperscript{1}$^*$
\quad
Xinliang Wang\textsuperscript{1}$^*$
\quad
Chunlong Xia\textsuperscript{1}$^*$\\
\quad
Shaojie Li\textsuperscript{1}$^*$
\quad
Shujie Yang\textsuperscript{1}
\quad
Teng Xi\textsuperscript{1}
\quad
Gang Zhang\textsuperscript{1}
\\[1em]
 \textsuperscript{1}Baidu Inc.
\quad
}

\maketitle

\begin{abstract}
With the continuous improvement of computing power and deep learning algorithms in recent years, the foundation model has grown in popularity. Because of its powerful capabilities and excellent performance, this technology is being adopted and applied by an increasing number of industries. In the intelligent transportation industry, artificial intelligence faces the following typical challenges: few shots, poor generalization, and a lack of multi-modal techniques. Foundation model technology can significantly alleviate the aforementioned issues. To address these, we designed the \textbf{1st Foundation Model Challenge}, with the goal of increasing the popularity of foundation model technology in traffic scenarios and promoting the rapid development of the intelligent transportation industry.
The challenge is divided into two tracks: all-in-one and cross-modal image retrieval. Furthermore, we provide a new baseline and benchmark for the two tracks, called Open-TransMind. According to our knowledge, Open-TransMind is the first open-source transportation foundation model with multi-task and multi-modal capabilities. Simultaneously, Open-TransMind can achieve state-of-the-art performance on detection, classification, and segmentation datasets of traffic scenarios. Our source code is available at \url{https://github.com/Traffic-X/Open-TransMind}.
\end{abstract}

\section{Introduction}
\label{sec:intro}

In recent years, the foundation model has gained a lot of attention in the artificial intelligence community and is one of the key areas for future advancement. Large-scale neural networks that have been trained on a massive amount of data and may be utilized to tackle complex problems are often the building blocks of foundation models. With deep learning's ongoing development, larger and larger deep learning models are being trained using an increasing amount of computing resources. This has sped up the development of foundation model technology, which is now finding numerous uses in areas like speech recognition, image and video processing, machine translation, and natural language processing.

\begin{figure}[t]
  \centering
  \includegraphics[width=1.0\linewidth]{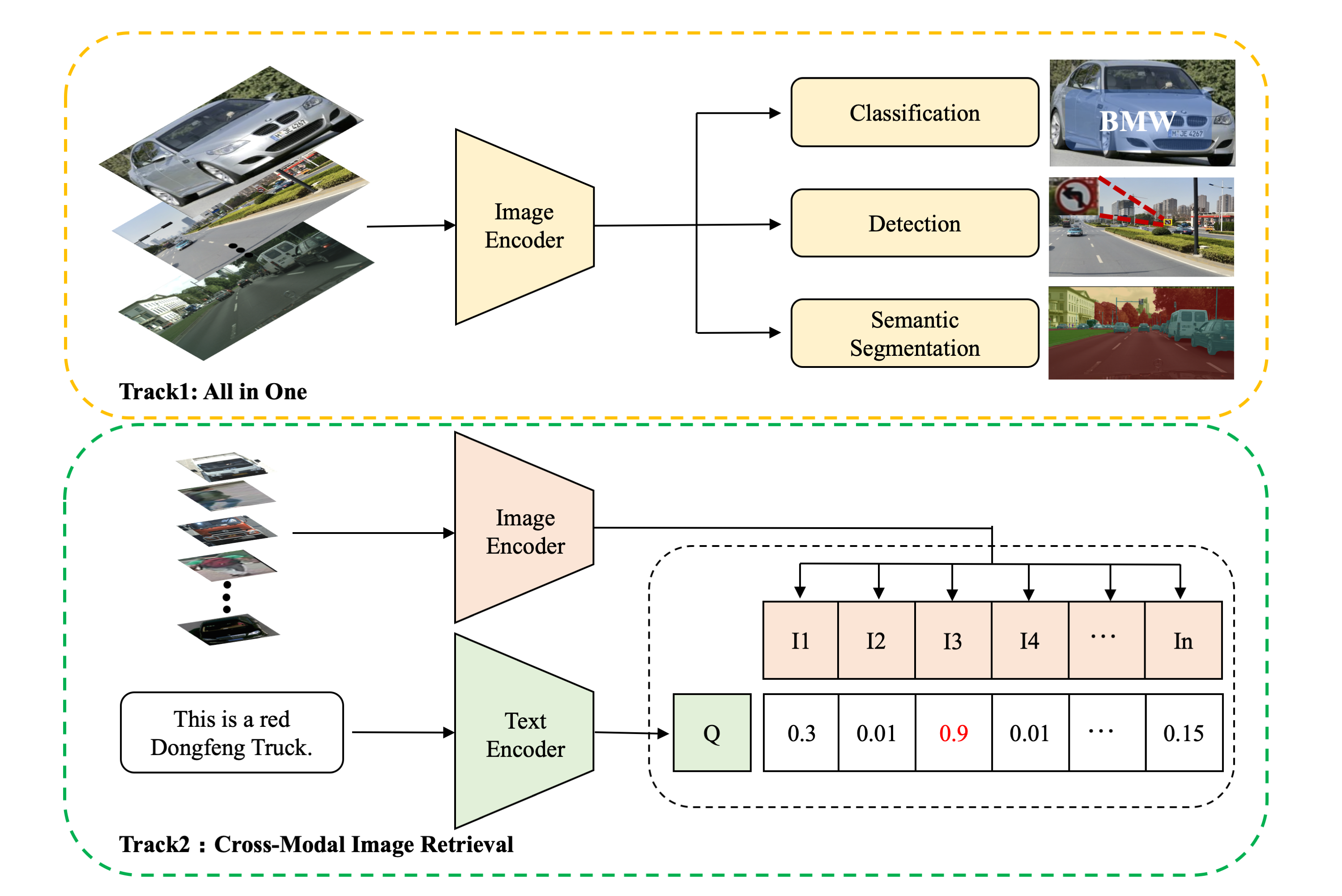}
   \caption{\textbf{Top:} Track 1 consists of  three tasks including classification, detection and semantic segmentation. \textbf{Bottom:} Track 2 is a text-image retrieval  task, where $Q$ represents the feature obtained by text encoding, and $I$ represents the images to be queried.}
   \label{fig:onecol}
\end{figure}

Compared to small-scale models, foundation models have many advantages. Foundation models can learn more from larger-scale datasets and perform better overall because of their higher computing power and more parameters. Furthermore, foundation models are capable of learning from data and adapting to new tasks, so they can better adapt to various application contexts. However, foundation models have some drawbacks, such as deployment and inference costs, so many works have conducted extensive research in this area ~\cite{xi2022ufo,cai2020once,Wang2020HATHT}.

Intelligent transportation is a field that integrates artificial intelligence technology and the transportation sector, focusing especially on autonomous driving, vehicle-infrastructure cooperation, traffic management, traffic information services, and other related topics. The following challenges face artificial intelligence in the field of intelligent transportation:

• Scenes with few shot problems have a weak effect;

• The data input, learning capacity, and generalization capacity of small-scale models are constrained;

• The pace of multi-modal technology development is slow, and industry applicability is restricted.

To address these issues, we designed the \textbf{1st Foundation Model Challenge} with the goal of increasing the popularity of foundation model technology in traffic scenarios. This activity can help strengthen the application of artificial intelligence technology in the transportation field and promote the rapid development of the intelligent transportation industry. The foundation model challenge consists of two tracks, as shown in Figure \ref{fig:onecol}:

\begin{table*}[t]
\centering
\begin{tabular}{c|c|ccc|cc}
\hline
\multirow{2}{*}{Task} & \multirow{2}{*}{Datasets}  & \multicolumn{3}{c|}{Perspective}                                          & \multicolumn{2}{c}{Modality}          \\ \cline{3-7} 
                      &                            & \multicolumn{1}{c|}{Roadside} & \multicolumn{1}{c|}{Vehicle} & Pedestrian & \multicolumn{1}{c|}{Visual} & Language \\ \hline
Classification        & Stanford Cars              & \multicolumn{1}{c|}{\checkmark}       & \multicolumn{1}{c|}{\checkmark}      & \checkmark         & \multicolumn{1}{c|}{\checkmark}     &          \\ 
Detection             & TT100K                     & \multicolumn{1}{c|}{\checkmark}       & \multicolumn{1}{c|}{}        & \checkmark         & \multicolumn{1}{c|}{\checkmark}     &          \\ 
Segmentation          & BDD100K                    & \multicolumn{1}{c|}{\checkmark}       & \multicolumn{1}{c|}{}        &            & \multicolumn{1}{c|}{\checkmark}     &          \\ 
Text-Image Retrieval  & PA100K+BIT-Vehicle+Web data & \multicolumn{1}{c|}{\checkmark}       & \multicolumn{1}{c|}{\checkmark}      & \checkmark         & \multicolumn{1}{c|}{\checkmark}     & \checkmark       \\ \hline
\end{tabular}
\caption{Overview of the used datasets for Track1 and Track2.}
\label{table:table2}
\end{table*}

\textbf{Track 1: All in One.} When designing well-crafted neural network structures and loss functions, multi-tasking can significantly improve model generalization. Using only one task's data for training may result in overfitting due to data noise in specific tasks. Unified large models will combine datasets from multiple tasks into unified training, allowing them to average noise from different jobs and thus improve their ability to learn better features. As a result, this track hopes that the unified foundation model will outperform multiple single-task models at the same time, allowing it to explore its upper bound on capabilities.

\textbf{Track 2: Cross-Modal Image Retrieval. }In traffic scenarios, high-performance image retrieval technology is critical for security management. The traditional image retrieval method typically begins with image attribute recognition and then achieves retrieval capability by comparing the image to the expected attribute. However, as multi-modal foundation model technology has advanced, text and image representation unification and modality transformation have become commonplace. The use of this capability can further improve the accuracy and flexibility of image retrieval. As a result, in this track, we created a text-image retrieval dataset that contestants can use to investigate multi-modal technology research, thereby improving the accuracy of text retrieval from images.


In this paper, we provide our baseline methodology, Open-TransMind, and a detailed benchmark introduction, including dataset compositions and evaluation strategies for these two tracks. Open-TransMind is, as far as we are aware, the first open-source foundation model for intelligent transportation and achieve state-of-the-art performance on detection, classification, and segmentation datasets of traffic scenarios. The open-source URL is \url{https://github.com/Traffic-X/Open-TransMind}. In the following chapters, we will introduce related work in Chapter 2, benchmark details, including dataset setup and evaluation strategies for each track, in Chapter 3, our baselines in Chapter 4, experiment analysis and results in Chapter 5, and the article conclusion in Chapter 6.





\section{Related Work}
\label{sec:formatting}
\subsection{Intelligent Transportation Public Datasets}
Publicly available datasets have played a critical role in advancing research in the field of intelligent transportation systems and autonomous driving by providing researchers with access to high-quality, diverse data to develop and test algorithms and systems. These datasets are divided into three task types: object detection, image classification, and semantic segmentation.

\textbf{Object Detection}. KITTI ~\cite{kitti} includes a large number of real-world scenes captured from moving vehicles, with data from various sensors such as cameras and LiDAR. Annotations of 2D and 3D bounding boxes for various types of objects, such as cars, pedestrians, and cyclists, are included in the dataset. TT100K ~\cite{tt100k} is a large-scale traffic sign detection benchmark comprised of 100,000 images of road scenes and 30,000 traffic sign instances each annotated with a class label, bounding box, and pixel mask. Other datasets, in addition to frontal view with sensors mounted on the vehicle, focus on roadside perception. Rope3D ~\cite{ye2022rope3d} creates designs for 3D object detection tasks in the roadside view. DAIR-V2X ~\cite{yu2022dair} is a 71,254 frame dataset of image and point cloud data designed for comprehensive research on 3D object detection in vehicle-infrastructure cooperative autonomous driving.

\textbf{Image Classification}. The PA100K dataset ~\cite{liu2017hydraplus} includes annotations for 26 different pedestrian attributes, such as gender, age group, clothing color, etc. The annotations are provided in the form of binary labels, indicating the presence or absence of each attribute in the corresponding pedestrian image. The Stanford Cars dataset ~\cite{KrauseStarkDengFei-Fei_3DRR2013} is a large-scale collection of car images collected for fine-grained classification. It includes 16,185 images of 196 different car models taken in a variety of settings and conditions.

\textbf{Semantic Segmentation}. The Cityscapes dataset ~\cite{Cordts2016Cityscapes} contains 5,000 high-resolution images captured in various weather and lighting conditions from 50 cities and provides pixel-level annotations for semantic segmentation tasks. Apolloscape ~\cite{ma2019trafficpredict} includes over 140,000 high-resolution images with 2D pixel-wise and 3D point-wise semantic annotation.

However, annotations for multiple tasks within a single image are not always present in the datasets mentioned above. The BDD100K dataset ~\cite{Yu2018BDD100KAD} provides annotations for multiple tasks, including object detection, semantic segmentation, lane detection, and so on, to aid research on multi-task learning in the field of autonomous driving and intelligent transportation systems.

\subsection{Foundation Models}
Large, pre-built neural network models that have been trained on massive amounts of data are referred to as pretrained foundation models. When applied to downstream tasks, foundational models can exhibit strong generalization capabilities with minimal customization. Foundation models are classified into three types based on their modality: LLMs (Large Language Models), LVMs (Large Vision Models), and VLMs (Large Vision-Language Models).

\textbf{LLMs.} Large language models like BERT ~\cite{devlin-etal-2019-bert} and GPT-3 ~\cite{NEURIPS2020_1457c0d6} have transformed natural language processing (NLP) by producing astounding results on a wide range of NLP tasks like language understanding, question answering, summarization, translation, and so on. LLMs usually adopt unsupervised learning techniques to pre-train transformer-based models ~\cite{NIPS2017_3f5ee243} on massive amounts of text data. 

\textbf{LVMs.} Previous large vision models were frequently based on contrastive learning on CNN-based architecture. However, given the enormous impact of transformer-based models in the field of NLP, researchers wonder whether they can apply the transformer concept to image data. ViT ~\cite{dosovitskiy2020vit} proposes modifying the transformer architecture to process images as a series of patches. Because of its large model capacity and generalizing capability, this transformer-based architecture is widely used in LVMs. MAE ~\cite{He_2022_CVPR} applies the concept of reconstruction from BERT to the domain of computer vision. UFO ~\cite{xi2022ufo} proposes a feature optimization paradigm that pretrains a large model on multiple tasks and then trims the model to a moderate size for deployment on specific tasks using NAS and Task-MoE.

\textbf{VLMs.} VLMs can perform both language and image-related tasks by combining LLMs and LVMs, such as image captioning, text-guided image generation, visual question answering, and so on. CLIP ~\cite{Radford2021LearningTV} is made up of a visual encoder and a text encoder, and it trains the model using contrastive learning on a large dataset of image-caption pairs. Following research has concentrated on the interaction of the image encoder and the text encoder. Some studies treat the two encoders as separate and independent models, while others ~\cite{pmlr-v139-kim21k} try to integrate them into a single, unified model. These studies seek to improve the model's performance by investigating various methods for combining and utilizing the visual and semantic features learned by the image and text encoders.



\section{Benchmark}
According to the perspective of data collection, the research content of intelligent transportation mainly includes vehicle perspective, roadside perspective, and pedestrian perspective scene tasks. These tasks include image classification, object detection, segmentation, and text-image retrieval, and the modalities include image, text, point cloud, etc. Considering the richness of tasks and resource occupation, we selected four tasks in three categories: roadside, vehicle, and pedestrian perspectives, which include image classification, object detection, semantic segmentation, and text-image retrieval (shown in Table \ref{table:table2}). In view of the above four tasks, we divide them into two tracks. Track 1 is composed of three tasks: image classification, object detection, and semantic segmentation. Track 2 is a cross-modal task of text-image retrieval. The following is a detailed benchmark description of Track 1 and Track 2.


\subsection{Track 1: All In One}
This track consists of three tasks: image classification, object detection, semantic segmentation. To accommodate as many perspectives and tasks as possible, we select three datasets: Stanford Car ~\cite{KrauseStarkDengFei-Fei_3DRR2013}, TT100K ~\cite{tt100k} and BDD100K ~\cite{Yu2018BDD100KAD}, which cover the vehicle, roadside, pedestrian perspective, and the three tasks listed above. Based on the chosen datasets, we created a method for multi-task co-evolution. We'll discuss datasets and evaluation measures in the sections below.

\textbf{Dataset.} 
BDD100K is a diverse visual-driven scene dataset gathered from vehicle perspectives that includes ten different tasks such as detection, segmentation, and tracking. As our benchmark, we chose the semantic segmentation task dataset with 720$\times$1280 resolution; the dataset contains geographic, environmental, and weather diversity, as well as 7,000 and 1,000 finely annotated images for training and testing, respectively.
Tsinghua-Tencent 100K (TT100K) is a large-scale traffic-sign detection and classification benchmark obtained from vehicle and pedestrian perspectives, with 100,000 images containing 30,000 traffic sign instances annotated with a class label, bounding box, and pixel mask. These images cover a wide range of lighting and weather conditions. We chose traffic-sign detection as our task, which included 6,107 and 3,067 annotated images for training and testing, respectively. Stanford Cars consists of 196 car classes with 360$\times$240 pixels, including 8,144 training images and 8,041 testing images. It is compiled from vehicle, roadside, and pedestrian perspectives. Brand, model, and year are used to categorize items.

\begin{table*}
\centering
\begin{tabular}{c|c|c|ccc|ccc}
\hline
\multirow{2}{*}{Method}         & \multirow{2}{*}{Backbone} & \multirow{2}{*}{Type}   & \multicolumn{3}{c|}{Tasks}                                                  & \multirow{2}{*}{\begin{tabular}[c]{@{}c@{}}Stanford Cars   \\ (acc\%)\end{tabular}} & \multirow{2}{*}{\begin{tabular}[c]{@{}c@{}}TT 100K   \\  (mAP\%)\end{tabular}} & \multirow{2}{*}{\begin{tabular}[c]{@{}c@{}}BDD100K   \\  (mIoU\%)\end{tabular}} \\ \cline{4-6}
                                &                           &                         & \multicolumn{1}{c|}{cls}       & \multicolumn{1}{c|}{det}       & seg       &                                                                                     &                                                                                &                                                                                 \\ \hline
CAP                             & xception                  & \multirow{3}{*}{single} & \multicolumn{1}{c|}{\checkmark} & \multicolumn{1}{c|}{}          &           & 95.70                                                                               & -                                                                              & -                                                                               \\
CABNet                          & vgg16                     &                         & \multicolumn{1}{c|}{}          & \multicolumn{1}{c|}{\checkmark} &           & -                                                                                   & 78.00                                                                          & -                                                                               \\
NiseNet                         & resnet101                 &                         & \multicolumn{1}{c|}{}          & \multicolumn{1}{c|}{}          & {\checkmark} & -                                                                                   & -                                                                              & 53.52                                                                           \\ \hline
\multirow{3}{*}{Open-TransMind} & \multirow{3}{*}{vit-base} & \multirow{3}{*}{single} & \multicolumn{1}{c|}{\checkmark} & \multicolumn{1}{c|}{}          &           & 82.78                                                                               & -                                                                              & -                                                                               \\
                                &                           &                         & \multicolumn{1}{c|}{}          & \multicolumn{1}{c|}{\checkmark} &           & -                                                                                   & 76.27                                                                          & -                                                                               \\
                                &                           &                         & \multicolumn{1}{c|}{}          & \multicolumn{1}{c|}{}          & {\checkmark} & -                                                                                   & -                                                                              & 52.55                                                                           \\ \hline
Open-TransMind                  & vit-base                  & multi                   & \multicolumn{1}{c|}{\checkmark} & \multicolumn{1}{c|}{\checkmark} & {\checkmark} & 91.64                                                                               & 76.90                                                                          & 55.13                                                                           \\ \hline
Open-TransMind                  & vit-base                  & task-moe               & \multicolumn{1}{c|}{\checkmark} & \multicolumn{1}{c|}{\checkmark} & {\checkmark} & 93.34                                                                               & 76.70                                                                          & 56.81                                                                           \\ \hline
\multirow{3}{*}{Open-TransMind} & \multirow{3}{*}{vit-huge} & \multirow{3}{*}{single} & \multicolumn{1}{c|}{\checkmark} & \multicolumn{1}{c|}{}          &           & 95.85                                                                               & -                                                                              & -                                                                               \\
                                &                           &                         & \multicolumn{1}{c|}{}          & \multicolumn{1}{c|}{\checkmark} &           & -                                                                                   & 82.10                                                                          & -                                                                               \\
                                &                           &                         & \multicolumn{1}{c|}{}          & \multicolumn{1}{c|}{}          & {\checkmark} & -                                                                                   & -                                                                              & 63.41                                                                           \\ \hline
Open-TransMind                  & vit-huge                  & multi                   & \multicolumn{1}{c|}{\checkmark} & \multicolumn{1}{c|}{\checkmark} & {\checkmark} & \textbf{95.96}                                                                      & \textbf{83.24}                                                                 & \textbf{64.80}                                                                  \\ \hline
\end{tabular}
\caption{Comparing our Open-TransMind approach based on the vit-huge backbone with other state-of-the-art methods and comparing the performance of single-task trained models and multi-task trained models based on the Open-TransMind with vit-base backbone on different task datasets.}
\label{table:all}
\end{table*}

\begin{figure}[t]
  \centering
  \includegraphics[width=1.0\linewidth]{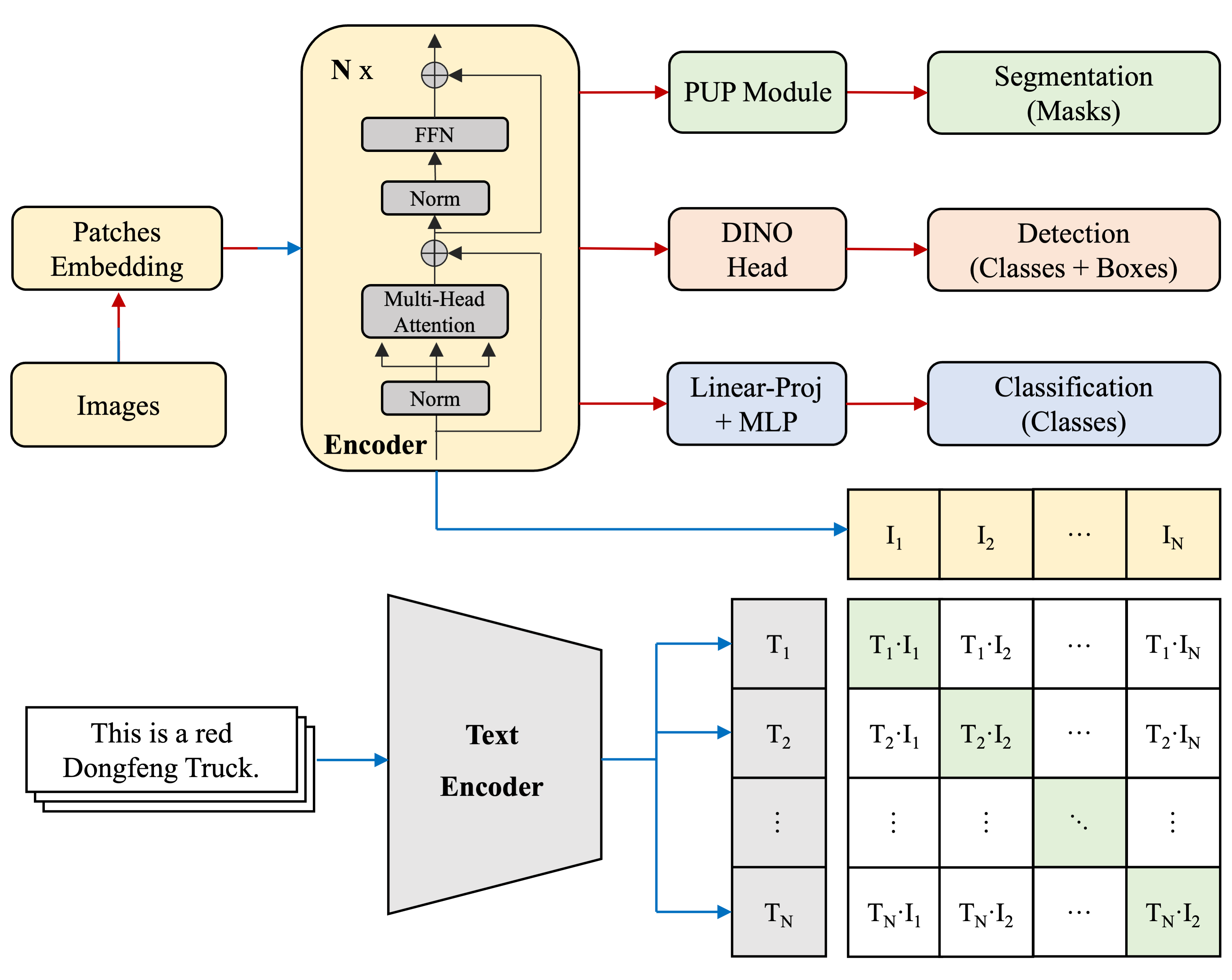}
   \caption{The framework of the large model. The red arrow represents the sub-framework of Track 1, and the blue represents the sub-framework of Track 2.}
   \label{fig:all-in-one}
\end{figure}

\textbf{Evaluation Metrics.}
This track hosts a collaborative multi-task optimization competition with classification, detection, and segmentation tasks. As a result, we include not only the evaluation metric for each task, but also a global metric that expresses the overall effect of the competition task. The classification's evaluation metric is Acc. mAP is the detection evaluation metric. The semantic segmentation evaluation metric is mIoU. As the global metric, this track uses the average of all metrics.

\subsection{Track 2: Cross-Modal Image Retrieval}
This track focuses on multi-modal text-image retrieval task involving two types of transportation participants: pedestrians and vehicles. We generate the images for this track using open-source datasets and internet data, and we used a large language model to generate text for each image. In terms of model design, we use a concise CLIP model and $mAP@k$ as the evaluation metric. Below is a detailed description of the data construction and evaluation metric calculation.

\begin{figure}[t]
  \centering
  \includegraphics[width=1.0\linewidth]{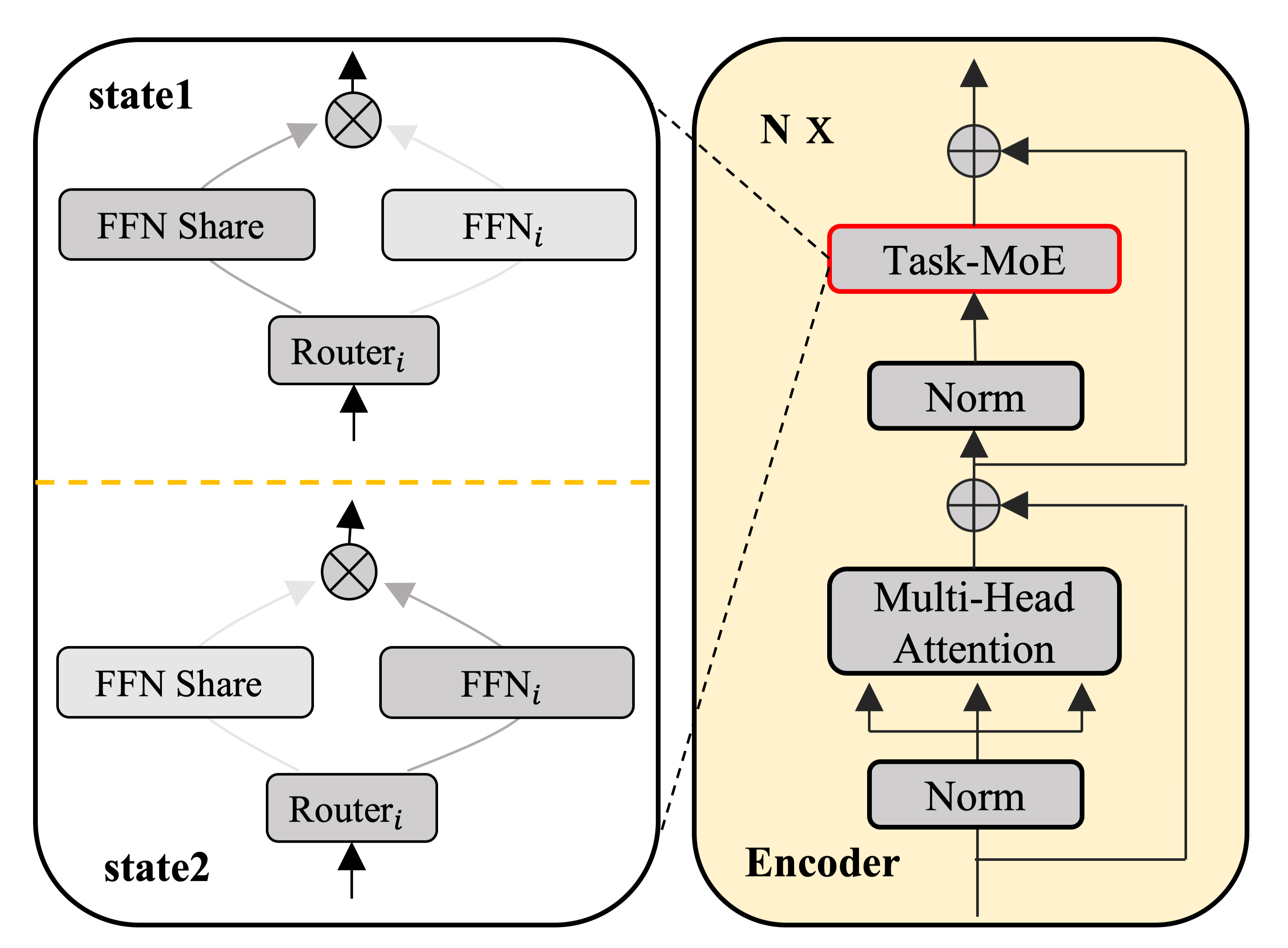}
   \caption{The structure of Task-MoE with shared FFN and specific FFN.}
   \label{fig:moe}
\end{figure}

\textbf{Dataset.} 
For pedestrian data, we use the open-source PA-100K \cite{liu2017hydraplus}, which is the largest dataset for pedestrian attributes recognition task. It contains 90,000 training images and 10,000 testing images with fine-grained attribute annotations such as pedestrian gender, age, upper and lower clothing style, whether or not they wear glasses and so on. Using a large language model, we generate descriptive text based on these attributes. As well as for vehicle data, we use an open-source dataset, BIT-Vehicle ~\cite{dong2015vehicle} to construct the test set. This dataset contains 9,850 vehicle images, with each image annotated with vehicle type attributes. Based on this dataset, we use a proprietary vehicle structured model that has a high accuracy and credibility  for vehicle color and brand classification (98.84\% for color and 96.48\% for brand), to annotate each image with additional vehicle color and brand attributes. After filtering out rare vehicle brand data and manually checking the accuracy of all attributes, we utilize a total of 7,611 images from the dataset, with each image annotated with vehicle color, brand, and type attributes. After statistical analysis, the test set we have constructed contains a total of 11 colors, 65 vehicle brands, and 6 vehicle types. We use these attributes as keywords to crawl 46,117 images from the internet as the training set, and generate corresponding descriptive texts for each image using a large language model based on the attribute category. It is important to note that we have not cleaned the web-scraped training set, which contains noise: certain images may have corresponding text with attribute errors.

After merging pedestrian and vehicle data, Track2 has a total of 137,117 data for training and 17,611 data for testing. Furthermore, in order to facilitate the validation of model capabilities, we randomly select 10,000 data from the training set as the validation set.

\textbf{Evaluation Metrics.}
The evaluation metric used in this competition is the mean Average Precision ($mAP@K$), where $K$ is set to 10. $mAP$ is a measure of the accuracy of text-based image retrieval. The calculation of $mAP@K$ is as follows:
$$mAP@K=\frac{1}{m} \times \sum_{i=1}^{K}{p(i)*\Delta r(i)}$$
where $m$ is the total number of text queries in the evaluation set, $p(i)$ represents the precision of the $topi$ retrieved results, and $\Delta r(i)$ is calculated as:
$$\Delta r(i)=r(i)-r(i-1)$$
where $r(i)$ is the recall of the $topi$ retrieved results and $r(0)=0$.

\section{Baseline}
In intelligent transportation scenarios, the types of tasks mainly involve multiple fields such as classification, detection, and segmentation. We can usually solve the corresponding problems by building a single-task model, but the generalization ability of the model will be poor due to the small size of the data. Considering the similarity between different tasks in the traffic scene, we propose a multi-task co-evolution model to promote mutual learning between different tasks, of course, there may be conflicts. In addition, traditional image retrieval methods usually use image attribute recognition to achieve retrieval capabilities. However, introducing cross-modal capabilities based on existing models, such as image-text retrieval, can further improve the flexibility of image retrieval applications. Therefore, we can use a multi-task co-evolutionary model to build an image model for text retrieval. Finally, we combine multi-task co-evolution and multi-modal text-image retrieval into a model called Open-TransMind.

\subsection{Task-MoE}
Our Open-TransMind model is built based on the UFO~\cite{xi2022ufo} multi-task and multi-path training framework. Due to the differences in the visual representation of the data used by different tasks, there may be conflicts between tasks. Further, we can alleviate conflicts by introducing Task-MoE. As shown in the Figure \ref{fig:moe}, the multi-path FFN module is set in the encoding module. Each task has two different path options, that is, to choose shared FFN (state1 in Figure \ref{fig:moe}) or specific FFN (state2 in Figure \ref{fig:moe}). All tasks will update the parameters of shared FFN, and specific tasks will only update the specific FFN parameters to achieve collaborative optimization between different tasks, and greatly improve the generalization ability of the model.

\subsection{Model Design for All in One}
The framework of the large multi-task model for this track is shown in Figure \ref{fig:all-in-one}. We select the backbone of the transformer architecture as the multi-task feature encoder. Based on this common feature, we respectively perform feature decoding for image classification, object detection, and semantic segmentation tasks and predict the results of corresponding tasks. The feature decoder for classification tasks consists of a linear-projection layer and a full-connection layer. The function of the linear-projection layer is to provide a layer of isolation between classification and multi-task features; the object detection decoder refers to the detection head of the DINO ~\cite{2022DINO} model, and the semantic segmentation decoder uses the progressive upsampling module of the SETR ~\cite{2021Rethinking} model.

\subsection{Model Design for Cross-Modal Image Retrieval}
The text-image retrieval model of this track is implemented using the same method as CLIP, and the overall framework is shown in the lower half of Figure \ref{fig:all-in-one}. We use the same transformer structure as Track 1 as the visual encoder, and use a text encoder to complete feature extraction of text data. During the training process, visual encoders and text encoders calculate visual and text features respectively, and achieve comparative learning of picture text pairs through contrastive training.

\section{Experiments}
\subsection{All In One}
\textbf{Implementation Details.} We chose vit-huge and vit-base models of transformers with different sizes as the backbone of Open-TransMind and set different output heads for the three different tasks of detection, segmentation, and classification. We merged three tasks and used a unified hyperparameter, such as the AdamW optimizer with an initial learning rate of 0.0001 and a weight decay of 0.0005. We adopt CosineAnnealingLR to drop the learning rate and implement different data augmentation methods for different tasks, including random resize, random cropping, random scaling in the range of 0.5 to 2.0, and random horizontal flipping, etc. Finally, we trained 120 epochs based on the ImageNet pre-training model with a batch size of 8 on eight A100 GPUs.

\begin{table}[t]
\centering
\begin{tabular}{c|cc}
\hline
Training Method                                  & Test & Val \\ \hline
training from scratch                      & 21.25 & 53.84  \\ 
CLIP                                    & 28.11 & 16.44  \\ 
CLIP finetune                           & 50.96 & 72.26  \\ 
CLIP finetune + All in One off-the-shelf & \textbf{53.54} & \textbf{72.56}  
\\ \hline
\end{tabular}
\caption{The quantitative results of the cross-modal text-image retrieval task on the test and validation sets we constructed.}
\label{table:track2_results}
\end{table}

\textbf{Evaluation Results.}
Based on the above experimental configuration, we conducted multiple sets of comparative ablation experiments. We chose the state-of-the-art methods described on \url{https://paperswithcode.com/} as baselines for comparison in order to confirm the efficacy of our approach, such as CAP, CABNet, and NiseNet, which are currently the best methods for classification, detection, and segmentation datasets of this track. To begin, we compared the performance of multi-task and single-task large models using Open-TransMind based on the vit-base backbone to validate the benefits of multi-task learning. As shown in Table \ref{table:all}, the multi-task joint training large model has significant advantages in the two tasks of segmentation and classification, which are respectively positive by 2.58\% and 8.86\%, and the detection task is slightly positive. At the same time, we compared the models trained using Task-MoE method and found that the classification and segmentation tasks performed better than multi-task joint training, while the detection task performance was basically the same, indicating that this method can alleviate conflicts between tasks. Multi-task joint training of large models has been shown to have higher generalization ability than single-task large models.
Second, we evaluated state-of-the-art methodologies against our Open-TransMind method built on the vit-huge backbone. Our vit-huge multi-task large model outperforms the vit-huge single-task large model and the vit-base large model, as can be observed. Meanwhile, it can achieve the state-of-the-art performance that is more advanced than baselines. Considering that large models need to occupy more hardware resources and take more training time, we provide Open-TransMind with vit-base backbone as our challenge code base.

\begin{figure}[t]
  \centering
  \includegraphics[width=1.0\linewidth]{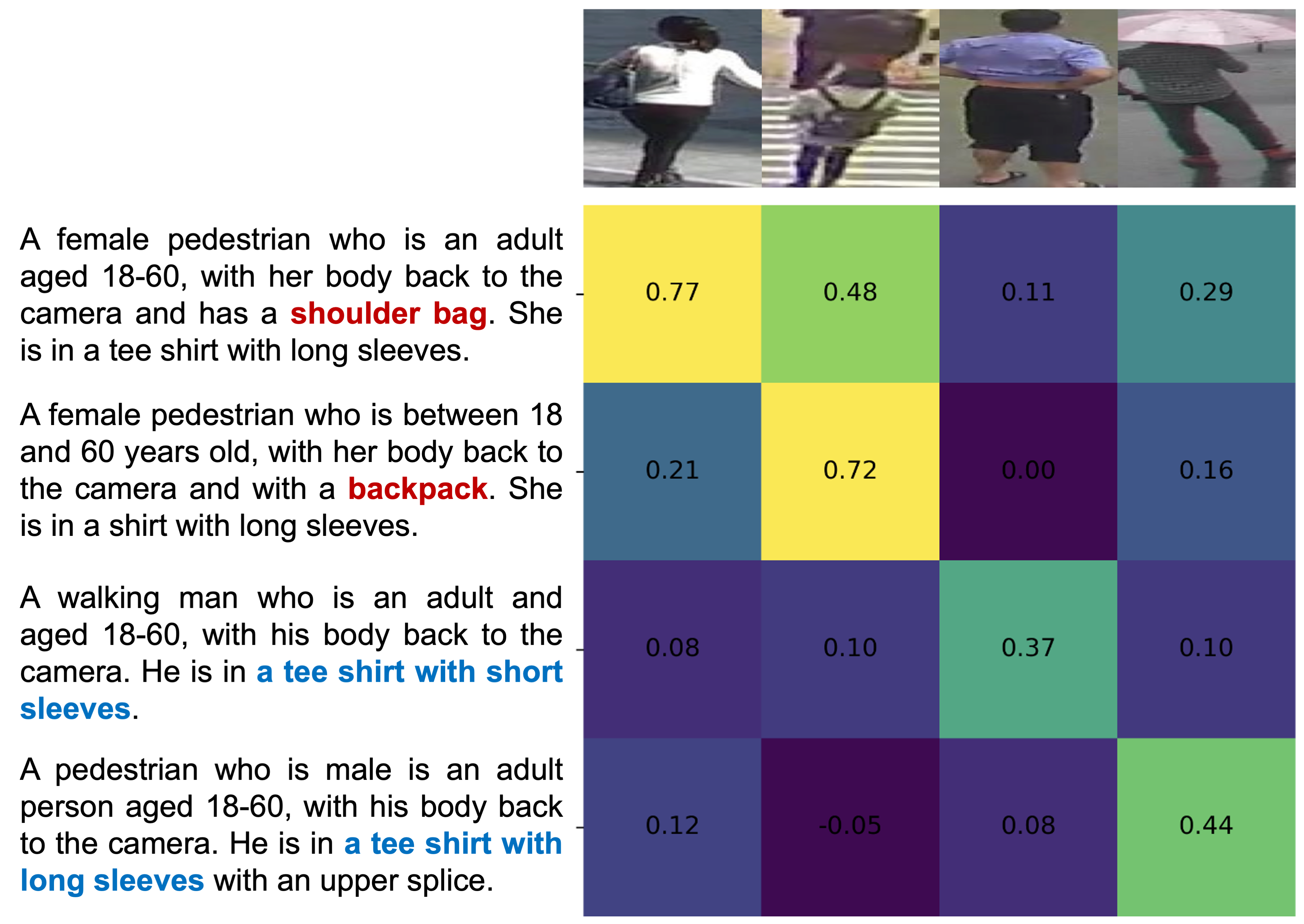}
   \caption{Visualization of similarity between different input texts and pedestrians with different attributes. Note that bright yellow represents high similarity, while dark blue represents low similarity.}
   \label{fig:retrieval_person}
\end{figure}

\subsection{Cross-Modal Image Retrieval}
\textbf{Implementation Details.} In the experiment, we use a transformer-based vit-base model as the visual feature encoder in the cross-modal Open-TransMind, and a naive transformer as the text feature encoder. We use an AdamW optimizer with an initial learning rate of 0.0001 and weight decay of 0.0005, and we adopt piecewise decay to reduce the learning rate. We only use basic resizing for data augmentation. The entire training process is conducted on 8 A100 GPUs for 20 epochs with a batch size of 128 for each GPU.

\textbf{Quantitative Results.} We conduct four comparative experiments: (1) training from scratch; (2) directly loading CLIP pre-training weights; (3) fine-tuning on the basis of CLIP pre-training weights; and (4) loading CLIP pre-training weights for fine-tuning and using the All-in-One model in Track 1 in an off-the-shelf manner, as an additional visual feature encoder to provide richer semantic information about the traffic scene.

As shown in Table \ref{table:track2_results} of the experimental results, we can see that the CLIP pre-training weights already have good multi-traffic participant text and image retrieval capabilities. After fine-tuning with real traffic scene data, CLIP achieves even better results. Among all the comparative experiments, the ``with All in One off-the-shelf'' method achieves the best results, which indicates that the All-in-One model has excellent traffic participant feature extraction ability through learning a large amount of traffic scene data.

\begin{figure}[t]
  \centering
  \includegraphics[width=1.0\linewidth]{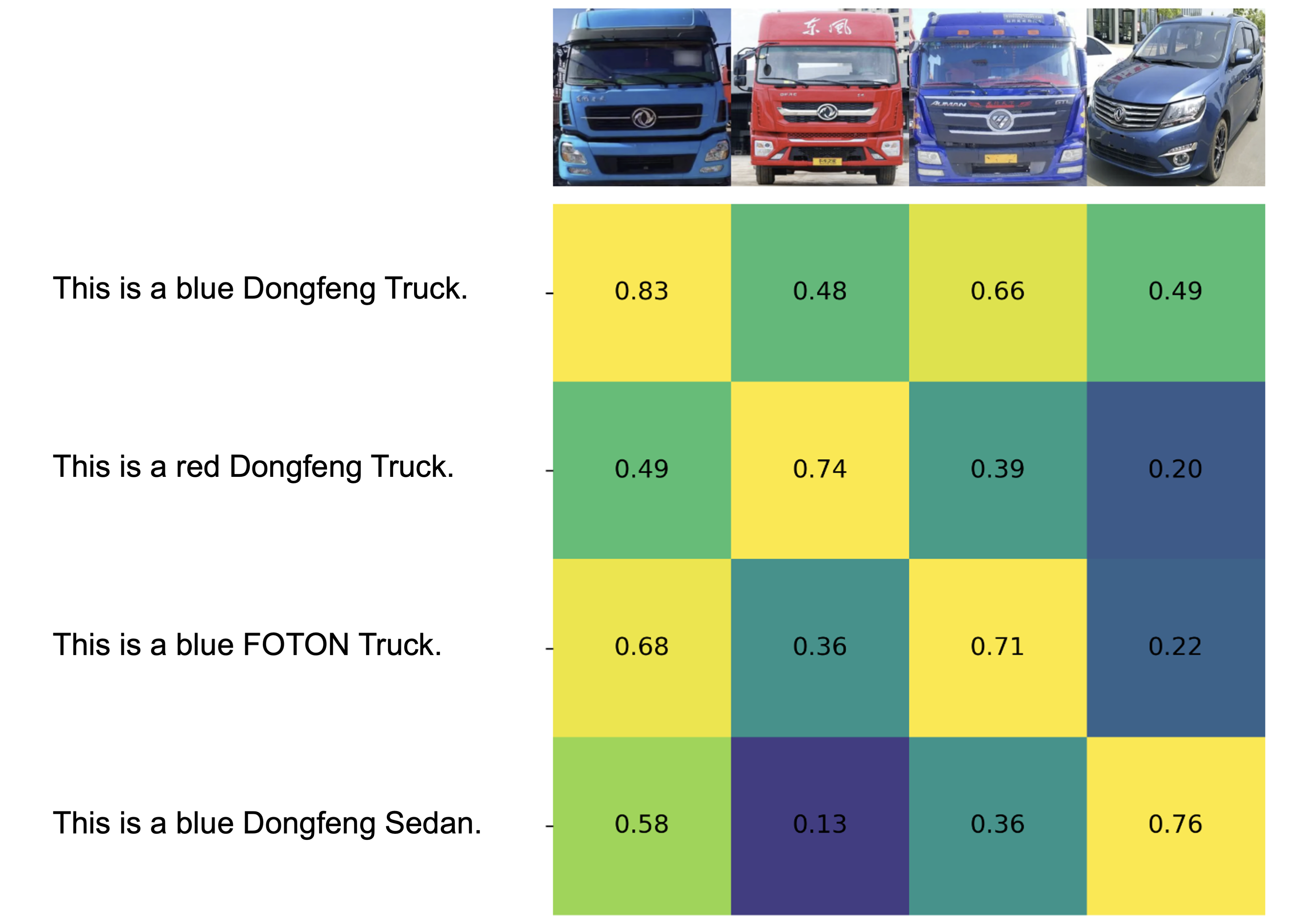}
   \caption{Visualization of similarity between different input texts and vehicles with different attributes. Note that bright yellow represents high similarity, while dark blue represents low similarity.}
   \label{fig:retrieval_car}
\end{figure}

\textbf{Qualitative Results.} Figure \ref{fig:retrieval_person} and Figure \ref{fig:retrieval_car} show the visualization of text-image retrieval, from which it can be seen that our model can effectively distinguish fine-grained differences in cross-modal information. In the pedestrian retrieval example in Figure \ref{fig:retrieval_person}, the query text in the first and second lines has the same attributes except for the different bag attributes. Open-TransMind can effectively distinguish this detailed attribute. The query text in the third and fourth lines has slight differences in the sleeve length of the shirt, and Open-TransMind can also provide correct retrieval results. In the vehicle retrieval example shown in Figure \ref{fig:retrieval_car}, the keywords in the query text in the first line are ``blue'', ``Dongfeng brand'', and ``truck''. Although the remaining three challenging query text changes the three attributes respectively, Open-TransMind can provide correct retrieval results. In addition, from this example, we can see that our model has certain knowledge emergence capabilities: there are three types of text and corresponding images in the training data: ``Blue Mercedes Benz'', ``Dongfeng brand'', and ``Truck'', but there is no text and corresponding images of ``Blue Dongfeng Truck''. The trained Open-TransMind can effectively complete the retrieval of ``Blue Dongfeng Truck'', as shown in the first column of row 1.

\section{Conclusion}
We propose a new baseline and benchmark for the first foundation model challenge of intelligent transportation, called Open-TransMind. Meanwhile, Open-TransMind is the first open-source foundation model of intelligent transportation with multi-task and multi-modal capabilities that we are aware of. Furthermore, Open-TransMind can achieve cutting-edge performance on traffic scenario detection, classification, and segmentation datasets, while also possessing cross-modal text-image retrieval capabilities. Contestants and researchers can investigate new solutions for multi-task co-evolution as well as the interrelationships of cross-modal. In the future, we plan to add more tasks, such as text recognition and point cloud detection. More researchers are expected to join the investigation of foundation model technology.
{\small
\bibliographystyle{ieee_fullname}
\bibliography{Open-TransMind}
}

\end{document}